\pdfoutput=1

\documentclass[11pt]{article}

\usepackage{acl}

\usepackage{times}
\usepackage{latexsym}
\usepackage{arydshln} 
\usepackage{graphicx} 

\usepackage[T1]{fontenc}

\usepackage[utf8]{inputenc}
\usepackage{booktabs}
\usepackage{multirow}
\usepackage{amsmath}
\usepackage{amsfonts}
\usepackage{bm}
\usepackage{microtype}

\title{Document-level Relation Extraction with Context Guided Mention Integration and Inter-pair Reasoning}
 \author{Chao Zhao$^{1,2}$, Daojian Zeng$^1$, Lu Xu$^1$, Jianhua Dai$^1$ \\
        \textsuperscript{\rm $1$}Hunan Normal University , Changsha, 410114, China \\ \textsuperscript{\rm $2$} Changsha University of Science \& Technology, Changsha, 410114, China\\ \texttt{zhaochao@stu.csust.edu.cn},  \texttt{zengdj@163.com},\texttt{\{xulu,daijianhua\}@hunnu.edu.cn}} 

\begin{document}
\maketitle
\begin{abstract}


Document-level Relation Extraction (DRE) aims to recognize the relations between two entities.
The entity may correspond to multiple mentions that span beyond sentence boundary.
Few previous studies have investigated the mention integration, which may be problematic because coreferential mentions do not equally contribute to a specific relation.
Moreover, prior efforts mainly focus on reasoning at entity-level rather than capturing the global interactions between entity pairs.
In this paper, we propose two novel techniques, \textbf{C}ontext \textbf{G}uided \textbf{M}ention \textbf{I}ntegration and \textbf{I}nter-pair \textbf{R}easoning (CGM2IR), to improve the DRE.
Instead of simply applying average pooling, the contexts are utilized to guide the integration of coreferential mentions in a weighted sum manner.
Additionally, inter-pair reasoning executes an iterative algorithm on the entity pair graph, so as to model the interdependency of relations.
We evaluate our CGM2IR model on three widely used benchmark datasets, namely DocRED, CDR, and GDA. 
Experimental results show that our model outperforms previous state-of-the-art models.
\end{abstract}

\section{Introduction}
\label{sec:intro}
Relation extraction is a fundamental problem in natural language processing, which aims to identify the semantic relation between a pair of entities mentioned in the text.
Recent progress in neural relation extraction has achieved great success \cite{zeng-etal-2015-distant,baldini-soares-etal-2019-matching}, but these approaches usually focus on binary relations (relations that only involve two entities) within a single sentence.
While in practice, a large number of relations in entity pairs span sentence boundaries\footnote{According to \citet{yao-etal-2019-docred}, at least 40.7\% of relations can only be identified from multiple sentences.}.
Many recent works \cite {yao-etal-2019-docred,DBLP:conf/aaai/Zhou0M021} pay emphasis on document-level scene that requires a larger context to identify relations, making it a more practical but also more challenging task.

\begin{figure}[t]
	\begin{center}
		\includegraphics[width=215pt]{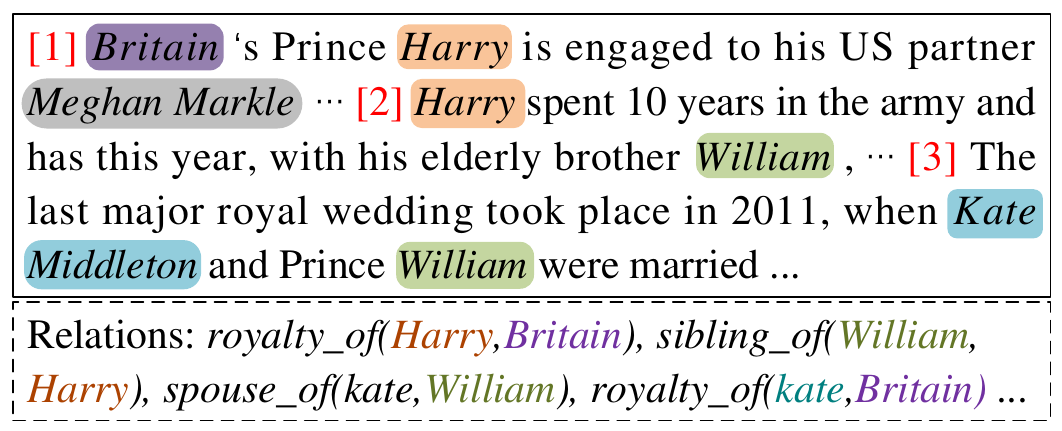}
		\caption{An example of DRE. Note that mentions of the same entity are marked with identical color.}
		\label{motivation}
	\end{center}
\end{figure}

Document-level Relation Extraction (DRE) poses unique challenges compared to its sentence-level counterpart.
First, it is more complex to model a document with rich entity structure for relation extraction.
The entities engaged in a relation may appear in different sentences, and some entities are repeated with the same phrases or aliases, the occurrences of which are often named entity mentions.
For example, as shown in Figure \ref{motivation}, \textit{Britain} and \textit{Kate} appear in the first and third sentences, respectively. 
\textit{Harry} and \textit{William} also appear more than once in this example.
We are therefore confronted to deal with cross-sentence dependencies and synthesizing the information of multiple mentions, in contrast to two entities in one sentence.
Second, there are intrinsic interactions among relational facts.
The identification of relations between two entities requires reasoning beyond the contextual features.
Specifically, in Figure \ref{motivation}, we can determine that the \textit{royalty\_of} relation exists between \textit{William} and \textit{Britain} from the context word \textit{Prince}. 
\textit{Kate} is also a member of the royal family, as she is married to \textit{William}.
Logical reasoning plays a dominant role when extract the fact $\left <Kate;royalty\_of;Britain\right>$.

Many previous works have tried to fulfill DRE and tackle the above challenges.
In order to exploit the document structure and capture cross-sentence dependencies, most current approaches construct a delicately designed document graph with syntactic structures (coreference, dependency, etc.) \cite{sahu-etal-2019-inter}, heuristics rules\footnote{For example, EoG \cite{christopoulou-etal-2019-connecting} builds a graph network with sentences, entities, and entity mentions as nodes and edges connected between different nodes according to heuristical rules.}, or structured attention \cite{nan-etal-2020-reasoning}.
The constructed graphs bridge entities that spread far apart in the document.
Besides, as Transformers for NLP can be considered as a graph neural network with multi-head attention as the neighbourhood aggregation function. It implicitly models long-distance dependencies.
There are also some works \cite{DBLP:conf/aaai/XuWLZM21,DBLP:conf/aaai/Zhou0M021} that attempt to use Pre-trained Models (PTMs) directly for DRE without involving graph structure.
Afterwards, researchers simply apply average (max) pooling to the representation of coreferential entity mentions.
Unfortunately, this is obviously not in accordance with intuition and fact.
All mentions are equally treated, ignoring the corresponding mention pair contexts for a specific relation.

In this paper, instead of simply synthesizing multiple coreferential mentions, we propose a novel context guided attention mechanism for mention integration.
Similarly to \citet{DBLP:conf/aaai/Zhou0M021}, after encoding through PTMs, we directly get the contexts for each entity pair from the attention heads.
Then, the contexts are guided as query to obtain the weights of mentions through cross-attention.
This process makes the representation of an entity change dynamically according to the entity pair in which it is located.

In light of the necessity of reasoning, message passing algorithms on graph are employed to update the entity representations accordingly.
Thus, it conducts reasoning in an implicit way \cite{christopoulou-etal-2019-connecting}. 
Otherwise, a special reasoning network is designed for relation inference \cite{zeng-etal-2020-double,li-etal-2021-mrn}.
Despite their great success, these methods mainly focus on entity-level or contextual information propagation rather than entity pair interactions, ignoring the global interdependency among multiple relational facts.

In this paper, we propose a novel inter-pair reasoning approach to achieve this purpose.
The head and tail entity representations obtained by context guided integration are merged with their contextual information to get the representations of multiple entity pairs.
Then, the entity pair representations are formed as the nodes of \textbf{G}raph \textbf{N}eural \textbf{N}etworks (GNNs).
The inter-pair interactions are captured through an iterative algorithm over entity pairs, so as to complete reasoning. 

By combining the proposed two techniques, we propose a simple yet effective document level relation extraction model, dubbed \textbf{CGM2IR} (\textbf{C}ontext \textbf{G}uided \textbf{M}ention \textbf{I}ntegration and \textbf{I}nter-pair \textbf{R}easoning), to fully utilize the power of PTMs.
To demonstrate the effectiveness of the proposed approach, we conduct comprehensive experiments on three widely used document level relation extraction datasets. 
The experimental results reveal that our CGM2IR model significantly outperforms the state-of-the-art methods.
Our contributions can be summarized as follows:
\begin{itemize}
	\item We propose a context guided attention mechanism to dynamically merge mentions that refer to the same entity in a weighted sum manner.
	Our approach innovatively uses contextual information to guide the entity representation.
	\item We propose an inter-pair reasoning approach to model interactions among entity pairs rather than entities.
	Reasoning based on entity pairs is more rational and consistent with the human way of intelligence and learning.
	\item We conduct experiments on three public DRE datasets.
	Experimental results demonstrate the effectiveness of our CGM2IR model that achieves the new state-of-the-art performance. 
\end{itemize}

\section{Related Work}

Relation extraction, also known as relational facts extraction, plays an essential role in a variety of applications in Natural Language Processing (NLP), especially for the automatic construction of Knowledge Graph (KG).
Early researchers mainly concentrate on the sentence-level task, i.e. predicting the relations between two entities within a sentence.
Many approaches \cite{zeng-etal-2014-relation,cai-etal-2016-bidirectional} have been proposed to effectively fulfill \textbf{S}entence-level \textbf{R}elation \textbf{E}xtraction (SRE), especially the pre-training-then-fine-tuning paradigm of PTMs \cite{zheng-etal-2021-prgc}.
SRE faces an inevitable restriction in practice, where many relation facts can only be extracted from multiple sentences.
Recently, researchers gradually push SRE forward to DRE \cite{yao-etal-2019-docred}.

In DRE, an entity may correspond to multiple mentions, which are scattered in different sentences.
We need to classify the relations of multiple entity pairs all at once, which usually requires complex reasoning skills and inter-sentential information.
DRE can be cast as a problem with multiple entity pairs to classify and multiple labels to assign \cite{DBLP:conf/aaai/Zhou0M021}.
To fulfill this task, most current approaches \cite{christopoulou-etal-2019-connecting} adopt appropriate models to first learn the contextual representation of an input document and encode the tokens in it. 
Then the representation of entity pairs is obtained by different strategies. 
Finally, a sigmoid classifier is used for multi-label classification.

\begin{figure*}[t]
	\begin{center}
		\includegraphics[width=400pt]{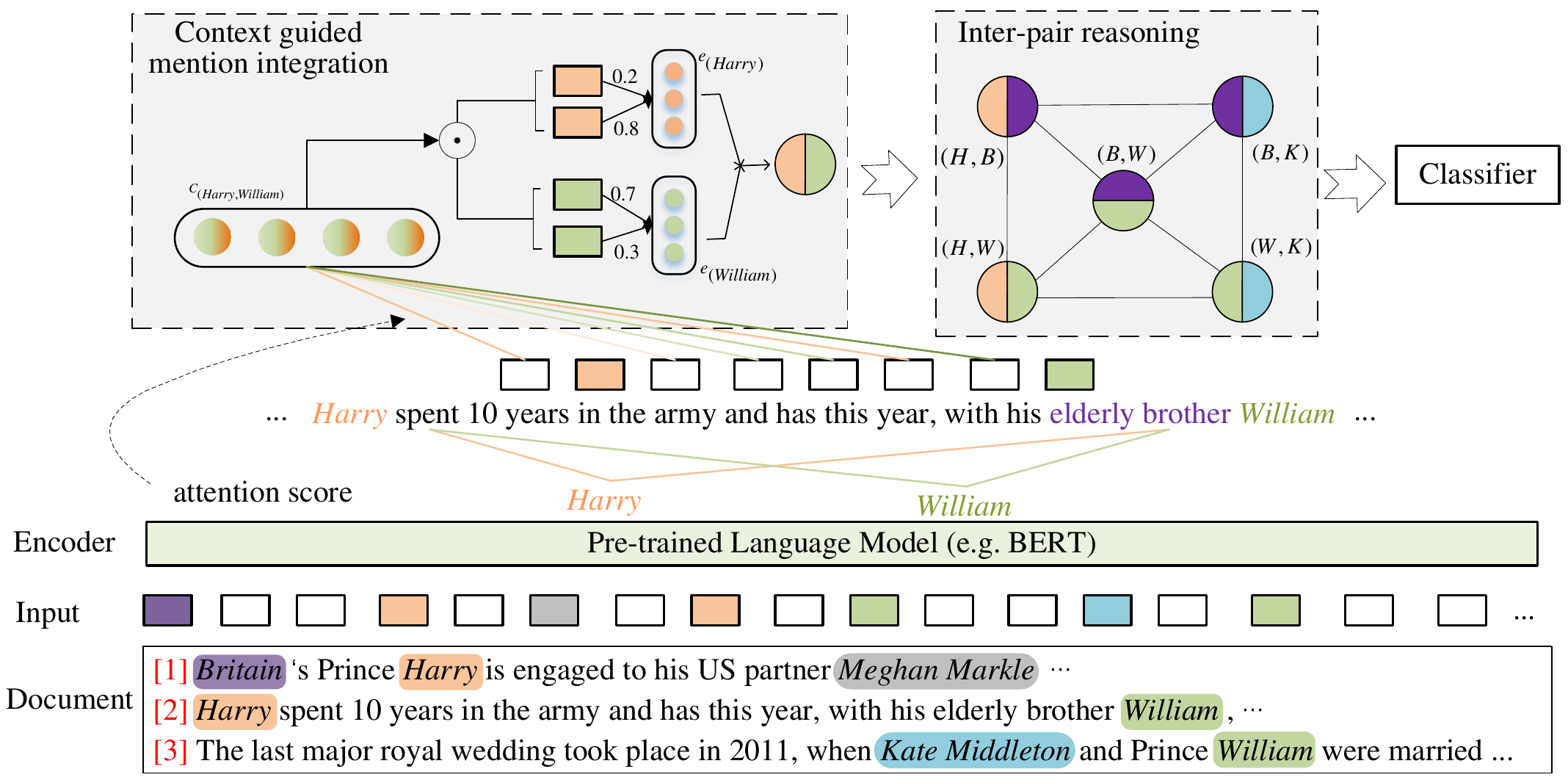}
		\caption{The overall architecture of CGM2IR. First, the input document is viewed as a long sequence of words, which are subsequently encoded through BERT. Then, the context guided mention integration module dynamically generates the head and tail entity embeddings for each entity pair. Next, we construct a homogeneous entity pair graph and use GNNs to model the inter-pair interaction. Finally, the classifier predicts relations of all the entity pairs in a parallel way. }
		\label{fig:overview}
	\end{center}
\end{figure*}

There are two types of mainstream for document representation.
On the one hand, researchers construct a delicately designed document graph.
\citet{quirk-poon-2017-distant} attempted a first step toward constructing document-level graph that augments conventional intra-sentential dependencies with new dependencies introduced by adjacent sentences and discourse relations.
Following this, \citet{christopoulou-etal-2019-connecting} built a document graph with heterogeneous types of nodes and edges.
\citet{nan-etal-2020-reasoning} proposed a latent structure induction to induce the dependency tree in the document dynamically.
\citet{wang-etal-2020-global}, \citet{zeng-etal-2020-double}, \citet{li-etal-2020-graph}, \citet{zhang-etal-2020-document} integrated similar structural dependencies to model documents.
Afterwards, graph based algorithm was employed to pass messages and conduct reasoning in an implicit way \cite{christopoulou-etal-2019-connecting}. 
Otherwise, a special reasoning network was designed for relation inference \cite{zeng-etal-2020-double,DBLP:conf/aaai/XuCZ21,li-etal-2021-mrn,xu-etal-2021-discriminative,zeng-etal-2021-sire}.
On the other hand, as Transformers for NLP can be considered as a graph neural network with multi-head attention as the neighbourhood aggregation function, it implicitly models long-distance dependencies.
There are also some works \cite{DBLP:journals/corr/abs-1909-11898, ye-etal-2020-coreferential} that attempt to use PTMs directly for DRE without involving graph structure.
\citet{DBLP:conf/aaai/XuWLZM21} incorporated entity structure dependencies within Transformers encoding part and throughout the overall system.
\citet{DBLP:conf/aaai/Zhou0M021} proposed an adaptive-thresholding loss and a localized context pooling to improve the performance.
Transformer-based approaches implicitly integrate reasoning into the encoding process.
These methods are simple but very effective, and have yielded the state-of-the-art performance.

Among the various amounts of prior works, \citet{DBLP:conf/aaai/Zhou0M021} and \citet{DBLP:conf/ijcai/ZhangCXDTCHSC21} are the most two relevant to our approach.
\citet{DBLP:conf/aaai/Zhou0M021} also considered context to enhance the entity representation.
\citet{DBLP:conf/ijcai/ZhangCXDTCHSC21} captured global interdependency among relation facts at entity pair level.
However, the differences are substantial.
First and foremost, these two approaches both equally treated all the mentions.
In contrast, we use a context guided intra-pair attention mechanism to weigh the mentions.
Moreover, we adopt a GNN that forms entity pairs as nodes to learn the inter-pair interactions.

\section{Methodology}
In this section, we describe the proposed model CGM2IR that incorporates context guided mention integration and inter-pair reasoning to improve DRE.
As illustrated in figure \ref{fig:overview}, CGM2IR mainly consists of four parts, namely (\romannumeral1) the document encoding; (\romannumeral2) the context guided mention integration; (\romannumeral3) the inter-pair reasoning; (\romannumeral4) and the final classification layer.

\subsection{Document Encoding Module}

To model the semantics of input document better, CGM2IR adopts BERT \cite{devlin-etal-2019-bert} as the document encoder, which has recently been proven surprisingly effective by presenting state-of-the-art results in various NLP tasks.

Given a document $D$ as input, it is comprised of $l$ tokens  $x=\{t_i\}^l _1$ and a set of annotated entities $e_i=\{m_j\}^t _1$ where entity $e_i$ may have multiple mentions that scatter across the document.
Borrowing the idea of entity marker \citep{baldini-soares-etal-2019-matching}, we first insert a special marker ``*'' at the start and end of mentions to mark the mention's span by the entity's annotation.
Then, the document encoder is responsible to map each token and mention markers to a sequence of contextualized embedding representations $\bm{H}=\{\bm{h_1}, \bm{h_2}, \cdots, \bm{h_n}\}$.
\begin{equation}
\bm{H} = PTMs(\{x_1, x_2, \cdots, x_n\})
\label{eq:encoder}
\end{equation}
where $n$ is the length of tokens with all markers.
For each mention, we take the embedding of start marker as the mention embeddings. 
Limited by the input length of BERT, we use a dynamic window \citep{DBLP:conf/aaai/Zhou0M021} to sequentially encode the whole documents when $n>512$.

\subsection{Context Guided Mention Integration Module}
\label{sec:context}
As argued in Section \ref{sec:intro}, an entity may be mentioned under the same phrase or alias in multiple sentences throughout the document.
To obtain entity-level representation, previous works usually synthesize the embeddings of all mentions of an entity.
These methods equally treat each mention and only generate one global embedding for an entity.
Then, the entity embedding is used in the relation classification of all entity pairs.

Unfortunately, it is obvious that some mentions may not be relevant to the relation when categorizing a particular entity pair.
Therefore, we propose a context guided attention mechanism that can generate fine-grained entity representations for each pair.
Different from the previous approaches, our motivation is to first get the entity-aware context through the average of mention attention matrices.
Then, the contexts involved in both head and tail entities are located to steer the model for mention integration.
Following \citet{DBLP:conf/aaai/Zhou0M021}, we explicitly use the token-level attention score $\bm{A}$ in the last encoder block of BERT to compute the pair-specific context embedding $\bm{c}_{h,t}$ for entity pair $(e_h, e_t)$ as follows:

\begin{equation}
\small
\begin{aligned}
\bm{c}_{(h,t)} &= \bm{H}\bm{a}_{(h,t)} \\
\bm{a}_{(h,t)} &= \frac{\bm{A}_h \cdot \bm{A}_t}{\bm{1}^\top(\bm{A}_h \cdot \bm{A}_t)} 
\end{aligned}
\label{eq:context_rep}
\end{equation}
where $\bm{A}_h = {avg_{m_i\in e_h}(A_{m_i})}$, $A_{m_i}$ is the attention matrix for $i$-th mention of head entity $e_h$ to all tokens in the document.
A similar operation yields $\bm{A}_t$ for the tail entity. 
Since the transformer-based PTMs have learned token-level dependencies well by training in a large-scale corpus, we attend all the tokens that are important to both entities in pair $(e_h, e_t)$ by multiplying their entity-level attentions score with a normalization.

After obtaining the contextual features of entity pairs in the first step, we use them as queries and perform cross-attention to pool the entity representations related to the entity pair from the mention embeddings of head or tail entity.
Specifically, given an entity pair $(e_h, e_t)$ and a sequence of mention embeddings $\bm{h}_{m_1},\bm{h}_{m_2},\cdots,\bm{h}_{m_p}$ of the head or tail entity, where $\bm{h}_{m_i}\in{R^d}$, $p$ is the number of mentions.
Guided by context feature $\bm{c}_{(h,t)} \in R^d$ for this pair, the head entity $\bm{e}^h _{(h,t)}$ is computed as follows:
\begin{equation}
\small
\begin{aligned}
\bm{e}^h _{(h,t)} &= \sum_{i=1}^{p} \bm{\alpha}^{i} _{(h,t)} \bm{h_{m_i}} \\
\bm{a}^{i} _{(h,t)} &= \frac{\bm{W}_Q\bm{c}^\top _{(h,t)} \bm{W}_K \bm{h_{m_i}}}{\sqrt{d}} \\
\bm{\alpha}^{i} _{(h,t)} &= \frac{\exp{(\bm{a}^{i} _{(h,t)})}} {\sum_{j=1}^{p}\exp{(\bm{a}^{j} _{(h,t)})}}
\end{aligned}
\label{eq:context_rep2}
\end{equation}
where $\bm{W}_Q\in\mathbb{R}^{d\times d}$, $\bm{W}_K\in\mathbb{R}^{d\times d}$ denotes the query and key transformation matrixes, $d$ is the dimension of hidden states.
In a similar way, we can obtain the representation of the tail entity $\bm{e}^t _{(h,t)}$.
We can observe that the representation of each entity is not fixed.
It is guided by the trade-off between the context and the entity pair in which it is located.
The head entity and the tail entity are combined to dynamically determine the respective representations.

\subsection{Inter-pair Reasoning Module}
To model interactions among entity pairs in a document, we construct a homogeneous entity pair graph and use GNNs to perform reasoning.

For each document $D$ with $m$ entities, we formulate a graph $\mathcal G = (\mathcal V, \mathcal E)$, where $m \times (m-1)$ entity pairs form the nodes of the graph.
Each node representation is computed by the following steps:
Given the embeddings $(\bm{e}^h _{(h,t)}, \bm{e}^t _{(h,t)})$ of an entity pair $(e_h, e_t)$ and its context features $\bm{c} _{(h,t)}$, we first combine the entity embeddings with their context embedding, and then map them to hidden representations $\bm{z}^{h} _{(h,t)}$ and $\bm{z}^{t} _{(h,t)}$ respectively with a feedforward neural network.
Finally, the entity pair embedding, $\bm{p}_{(h,t)}$, is calculated through a group bilinear\footnote{Group bilinear \citep{DBLP:conf/nips/ZhengFZL19} splits the embedding dimensions into $k$ equal-sized groups and applies bilinear within the groups.} function as follows:
\begin{equation}
\small
\begin{aligned}
\bm{z}^{h} _{(h,t)} &= tanh(W_h\bm{e}^h _{(h,t)}+W_{c_1}\bm{c}_{(h,t)}) \\
\bm{z}^{t} _{(h,t)} &= tanh(W_t\bm{e}^t _{(h,t)}+W_{c_2}\bm{c}_{(h,t)}) \\
\bm{p}_{(h,t)} &= \sigma(\sum_{i=1}^{k} \bm{z}^{h^i \top} _{(h,t)} W^i _p  \bm{z}^{t^i} _{(h,t)})
\end{aligned}
\label{eq:pair1}
\end{equation}
where $W_h\in\mathbb{R}^{d\times d}$, $W_t\in\mathbb{R}^{d\times d}$, $W_{c_1}\in\mathbb{R}^{d\times d}$, $W_{c_2}\in\mathbb{R}^{d\times d}$ and $ W^i _p\in\mathbb{R}^{d/k\times d/k}$ are learnable parameters. 
Furthermore, we concatenate the entity pair embedding with coreference embedding for head and tail entities to get the initial node representations following \citet{yao-etal-2019-docred}:

\begin{equation}
\small
\bm{P}^0 _{(h,t)} = [\bm{p^h};\bm{p} _{(h,t)};\bm{p^t}]
\label{eq:pair2}
\end{equation}
In contrast to the fully-connected case, we link each node to the nodes that have overlapping entities with it, since the clues for logical reasoning are usually passed on the chain of entities as it is approved in \citet{xu-etal-2021-discriminative} and \citet{zeng-etal-2021-sire}.
After the graph is constructed, We use GNNs to learn the inter-pair interactions.
In each layer $l$, The GNNs selectively aggregate all entity pair embeddings passed from neighbors through an attention mechanism to update its representation in the next layer $l+1$.
Formally, we have:
\begin{equation}
\small
\begin{aligned}
\bm{P}^{l+1} _u &= FFN(\bm{W}_r\sum_{\bm{v} \in \mathcal{N}_{(u)}}\alpha_{(u,v)}\bm{P}^l_u) \\
\alpha_{(u,v)} &= \frac{\exp[{\bm{Q}\bm{P}^l _v(\bm{K}\bm{P}^l _u)^\top}]}  {\sum_{\bm{v^\prime} \in \mathcal{N}_{(u)}} \exp[{\bm{Q}\bm{P}^l _{v^\prime}(\bm{K}\bm{P}^l _u)^\top}]} 
\end{aligned}
\label{eq:interpair3}
\end{equation}
where $\bm{W}_r\in\mathbb{R}^{d\times d}$, $\bm{Q}\in\mathbb{R}^{d\times d}$, $\bm{K}\in\mathbb{R}^{d\times d}$ are learnable weight matrices, $FFN(·)$ denotes a feed-forward network, $\mathcal{N}_{(u)}$ is the set of neighbor nodes to the vertex $u$.
In addition, we employ residual connection between two layers and perform layer normalization.
\subsection{Classification Module}

To determine the semantic relations for an entity pair $(e_h, e_t)$, we first concatenate two pair-specific entity representations and the corresponding final entity pair representation.
\begin{equation}
\small
\bm{r}_{(h,t)} = [\bm{e}^{h} _{(h,t)};\bm{e}^{t} _{(h,t)};\bm{P} _{(h,t)}]
\label{eq:classification}
\end{equation}
Then, we use a feed-forward neural network to calculate the probability for each relation:
\begin{equation}
\small
\bm{P}(r|e_h,e_t) = sigmoid(\bm{W}_b\sigma(\bm{W}_a\bm{r}_{(h,t)} + \bm{b}_a) + \bm{b}_b)
\label{eq:classification}
\end{equation}
where $\bm{W}_a\in\mathbb{R}^{3d\times d}$, $\bm{W}_b\in\mathbb{R}^{d\times r}$, $\bm{b}_a$, $\bm{b}_b$ are learnable parameters, $\sigma$ is an elementwise activation function (e.g., tanh).

To address the multi-label and sample imbalance problem more effectively, we adopt an adaptive-thresholding loss \citep{DBLP:conf/aaai/Zhou0M021} as the classification loss to train our model in an end-to-end way.
Specifically, it introduces an additional threshold relation category \textit{TH}, and optimizes the loss by increasing the logits of the positive relations $\mathcal{P}_T$ higher than the \textit{TH} relation and decreasing the logits of the negative relations $\mathcal{N}_T$ lower than the \textit{TH} relation.
\begin{equation}
\begin{aligned}
\small
& \mathcal L = -\sum_{r \in \mathcal{P}_T}log(\frac{ \exp(logit_r)) } {\sum_{r^\prime \in \mathcal{P}_T\cup\{TH\}} \exp(logit_r) }) \\ &- log(\frac{ \exp(logit_{TH})) } {\sum_{r^\prime \in \mathcal{N}_T\cup\{TH\}} \exp(logit_r) }) \hfill
\end{aligned}
\label{lossre}
\end{equation}
where $logit$ is the output in the last layer before Sigmoid function.

\section{Experiments}
\subsection{Datasets}

We evaluate the effectiveness of our CGM2IR model on three public DRE datasets: DocRED, CDR, and GDA.
The dataset statistics are shown in Table \ref{tab:statistics}.

\textbf{DocRED} is a large-scale human-annotated dataset for document-level RE proposed by \citep{yao-etal-2019-docred}.
It contains 97 types of relations and 5,053 annotated documents in total which are constructed from Wikipedia and Wikidata.
Documents in DocRED contain about 12.6 positive relational facts on average, which is several times that of the common sentence-level RE dataset.
\textbf{CDR} (Chemical-Disease Reactions) \citep{DBLP:journals/biodb/LiSJSWLDMWL16} and \textbf{GDA} (Gene-Disease Associations) \citep{DBLP:conf/recomb/WuLLTL19} are two widely-used DRE datasets in the biomedical domain.
They both contain only one type of positive relation, \textit{Chemical-Induced-Disease} between chemical and disease entities and \textit{Gene-Induced-Disease} between gene and disease entities respectively.
For a fair comparison, We follow the standard split of the three datasets as \citet{zeng-etal-2020-double} and \citet{DBLP:conf/aaai/Zhou0M021}.

\begin{table}[t]
	\small
	\centering
	\begin{tabular}{lccc}
		\toprule
		Statistics / Dataset & DocRED &  CDR & GDA \\
		\midrule
		\# Train & 3,053 &  500 & 23,353 \\
		\# Dev & 1,000 &  500 & 5,839 \\
		\# Test & 1,000 &  500 & 1,000 \\
		\# Relations & 97 &  2 & 2 \\
		Avg. \# Ment. per Ent. & 1.4 &  2.7 & 3.3 \\
		Avg. \# Ents per Doc. & 19.5 &  7.6 & 2.4 \\
		Avg. \# Facts. per Doc. & 12.6 &  2.1 & 1.5 \\
		\bottomrule
	\end{tabular}
	\caption{Statistics of the datasets. }
	\label{tab:statistics}
\end{table}
\subsection{Experiment Settings and Evaluation Metrics}
\begin{table*}[h]
	\centering
	\small
	\begin{tabular}{lcccc}
		\toprule
		\multirow{2}{*}{\textbf{Model}} & \multicolumn{2}{c}{\textbf{Dev}} & \multicolumn{2}{c}{\textbf{Test}} \\ \cline{2-5} 
		& Ign $F_1$  &  $F_1$  	& Ign $F_1$  &  $F_1$ \\ \hline 
		LSR-BERT$_{base}$ \citep{nan-etal-2020-reasoning} & 52.43     & 59.00  & 56.97  & 59.05      \\ 
		GEDA-BERT$_{base}$ \citep{li-etal-2020-graph} & 54.52 &  56.16 &  53.71  & 55.74      \\
		GCGCN-BERT$_{base}$ \citep{zhou-etal-2020-global} & 55.43  & 57.35 & 54.53    & 56.67  \\
		GLRE-BERT$_{base}$ \citep{wang-etal-2020-global} & -     & -  & 55.40    & 57.40  \\
		HeterGSAN-BERT$_{base}$ \citep{DBLP:conf/aaai/XuCZ21} & 58.13 &  60.18 &  57.12 &  59.45  \\	
		GAIN-BERT$_{base}$ \citep{zeng-etal-2020-double} & 59.14 & 61.22  &  59.00 &  61.24  \\ 
		DRE-BERT$_{base}$ \citep{xu-etal-2021-discriminative} & 59.33 &  61.39 &  59.15  & 61.37  \\
		SIRE-BERT$_{base}$ \citep{zeng-etal-2021-sire} &  59.82 &  61.60   & 60.18 &  62.05  \\  \midrule \midrule
		BERT$_{base}$ \citep{DBLP:journals/corr/abs-1909-11898} & - &  54.16 &  - &  53.20  \\ 
		HIN-BERT$_{base}$ \citep{DBLP:conf/pakdd/TangC0CFWY20} & 54.29  & 56.31 &  53.70  & 55.60  \\ 
		CorefBERT$_{base}$ \citep{ye-etal-2020-coreferential} & 55.32 &  57.51 & 54.54 & 56.96  \\ 
		SSAN-BERT$_{base}$ \citep{DBLP:conf/aaai/XuWLZM21} & 57.03 &  59.19 & 55.84 & 58.16  \\ 
		ATLOP-BERT$_{base}$ \citep{DBLP:conf/aaai/Zhou0M021} & 59.22 &  61.09 & 59.31 & 61.30  \\ 
		MRN-BERT$_{base}$ \citep{li-etal-2021-mrn} & 59.74 &  61.61  & 59.52 & 61.74  \\ 
		DocuNet-BERT$_{base}$\citep{DBLP:conf/ijcai/ZhangCXDTCHSC21} & 59.86     & 61.83  & 59.93   & 61.86  \\ \hdashline 
		\textbf{CGM2IR-BERT$_{base}$}  & \textbf{60.02}  & \textbf{62.01}  & \textbf{60.24}  & \textbf{62.06}  \\ \hline 
		BERT$_{large}$ \citep{DBLP:journals/corr/abs-1909-11898} & 56.67 & 58.83 & 56.47 & 58.69  \\ 
		CorefRoBERTa$_{large}$ \citep{ye-etal-2020-coreferential} & 57.84 & 59.93 &  57.68 & 59.91  \\ 
		SSAN-RoBERTa$_{large}$ \citep{DBLP:conf/aaai/XuWLZM21} & 60.25 &  62.08  & 59.47 & 61.42  \\ 
		GAIN-BERT$_{large}$ \citep{zeng-etal-2020-double} & 60.87 & 63.09  &  60.31  &  62.76  \\ 
		ATLOP-RoBERTa$_{large}$ \citep{DBLP:conf/aaai/Zhou0M021} & 61.32 &  63.18 &  61.39 & 63.40  \\ \hdashline
		\textbf{CGM2IR-RoBERTa$_{large}$}  & \textbf{62.03}     & \textbf{63.95}  & \textbf{61.96}  & \textbf{63.89}  \\ \hline \bottomrule

	\end{tabular}
	\caption{Results on the development and test set of DocRED. We separate graph-based and non-graph-based methods into two groups. The results of baselines are from their related papers.}
	\label{tab:result_docred}
\end{table*}
In our CGM2IR implementation, we use cased BERT-base \citep{devlin-etal-2019-bert} or RoBERTa-large \citep{DBLP:journals/corr/abs-1907-11692} the encoder on DocRED and cased SciBERT-base \citep{beltagy-etal-2019-scibert} on CDR and GDA. 
AdamW \citep{DBLP:conf/iclr/LoshchilovH19} is used to optimize the neural networks with a linear warmup and cosine decay learning rate schedule.
We set the initial learning rate for all encoder modules to $2e^{-5}$, the initial learning rate for other modules to $1e^{-4}$, the embedding dimension, and the hidden dimension to 768.
The GNNs have 3 layers and the hidden size of node embedding is 768. 
All hyper-parameters are tuned based on the development set.
Other parameters in the network are all obtained by random orthogonal initialization \citep{DBLP:journals/corr/SaxeMG13} and updated during training.
All the experiments are trained with an NVIDIA RTX 3090 GPU.

Following \citet{yao-etal-2019-docred} and previous works, we use the micro $F_1$ and micro Ign $F_1$ as the evaluation metrics for DocRED. 
Ign $F_1$  denotes the result after excluding the common relational facts that appear in both training set and development/test sets.
For CDR and GDA, in addition to using micro $F_1$, we also report the  Intra $F_1$ and Inter $F_1$ metrics to evaluate the model's performance on intra-sentential relations and inter-sentential relations on the dev set, since they strictly annotate these two types of facts but DocRED does not.
In our experiments, a triplet is taken as correct when the two corresponding entities and the relation type are all correct and we exclude all triplets with relation of ``None''.

\subsection{Results on DocRED}

We conduct comprehensive and comparable experiments on DocRED dataset. 
The results are shown in Table \ref{tab:result_docred}.

We compare our CGM2IR model with lots of methods from two categories.
The first one is graph-based methods, including LSR \citep{nan-etal-2020-reasoning}, GEDA \citep{li-etal-2020-graph}, GCGCN-BERT \citep{zhou-etal-2020-global}, GLRE \citep{wang-etal-2020-global}, GAIN \citep{zeng-etal-2020-double}, HeterGSAN \citep{DBLP:conf/aaai/XuCZ21}, SIRE \citep{zeng-etal-2021-sire} and DRE \citep{xu-etal-2021-discriminative}.
The second one is non-graph-based methods including BERT \citep{DBLP:journals/corr/abs-1909-11898}, HIN-BERT \citep{DBLP:conf/pakdd/TangC0CFWY20}, CorefBERT \citep{ye-etal-2020-coreferential}, SSAN \citep{DBLP:conf/aaai/XuWLZM21}, ATLOP \citep{DBLP:conf/aaai/Zhou0M021}, MRN \citep{li-etal-2021-mrn} and DocuNet \citep{DBLP:conf/ijcai/ZhangCXDTCHSC21}. 
The baselines we selected all use BERT as their encoder.

As shown in Table \ref{tab:result_docred}, we observe that CGM2IR outperforms all baseline methods on both development and test sets.
Compared with the models in these two categories, both $F_1$ and Ign $F_1$ of our model are significantly improved.
Among the various amounts of baselines, ATLOP \citep{DBLP:conf/aaai/Zhou0M021} and DocuNet \cite{DBLP:conf/ijcai/ZhangCXDTCHSC21} are the most two relevant to our approach.
Compared to ATLOP-BERT$_{base}$, the performance of CGM2IR-BERT$_{base}$ improves roughly about $0.8\%$ for Ign $F_1$ and $0.92\%$ for $F_1$.
CGM2IR-BERT$_{base}$ also brings about $0.2\%$ ign $F_1$ enhancement compared to DocuNet-BERT$_{base}$, which verifies the effectiveness of our proposed method.
Furthermore, CGM2IR-RoBERTa$_{large}$ obtains better results than baselines with BERT-large or RoBERTa-large as well.
For example, CGM2IR-RoBERTa$_{large}$ achieves $0.71\%$ Ign $F_1$/$0.77\%$ $F_1$ gain compared to ATLOP-RoBERTa$_{large}$ on the development set. 
In general, these results demonstrate both the effectiveness of context guided mention integration and the usefulness of inter-pair reasoning.
\begin{table}[t]
	\centering
	\small
	\begin{tabular}{lcccc}
		\toprule
		Model &   $F_1$ &  intra-$F_1$ &  inter-$F_1$\\
		\midrule
		\textbf{$\bullet$ CDR} Dataset & & &\\ \hdashline
		EoG  &  63.6 & 68.2 & 50.9  \\ 
		LSR  & 64.8  &68.9  &53.1 \\ 
		DHG-BERT$_{base}$  &   65.9  & 70.1   &  54.6 \\ 
		MRN  & 65.9 & 70.4   & 54.2 \\  
		ATLOP-SciBERT$_{base}$   &  69.2 & 74.2 & 52.6 \\ \hdashline
		\textbf{CGM2IR-SciBERT$_{base}$}   & \textbf{73.8} &\textbf{ 79.2} &\textbf{55.1}  \\  \hline \hline
		\textbf{$\bullet$ GDA} Dataset & & &\\ \hdashline
		EoG  &  81.5 & 85.2 &  50.0  \\ 
		LSR  &  82.2 & 85.4  & 51.1 \\ 
		MRN  & 82.9  & 86.1  & 53.5 \\  
		DHG-BERT$_{base}$  &  83.1 & 85.6 & 58.8  \\ 
		ATLOP-SciBERT$_{base}$   &  83.9 & 87.3  & 52.9  \\ \hdashline
		\textbf{CGM2IR-SciBERT$_{base}$}    & \textbf{84.7}  & \textbf{88.3}  & \textbf{59.0}  \\
		\bottomrule
	\end{tabular}
	\caption{Results on CDR and GDA datasets.}
	\label{tab:result_cdrgda}
\end{table}
\subsection{Results on CDR and GDA}
Table \ref*{tab:result_cdrgda} depicts the comparisons with state-of-the-art models on CDR and GDA.
We compare our CGM2IR model with five baselines, including EoG \citep{christopoulou-etal-2019-connecting}, DHG \citep{zhang-etal-2020-document}, LSR \citep{nan-etal-2020-reasoning}, MRN \citep{li-etal-2021-mrn}, ATLOP \citep{DBLP:conf/aaai/Zhou0M021}.
Our model adopts SciBERT$_{base}$ for its superiority when dealing with biomedical domain texts.

It can be observed that CGM2IR achieves the new state-of-the-art $F_1$ score on these two datasets in the biomedical domain.
On CDR test set, CGM2IR obtains +4.6 $F_1$ gain, which significantly outperforms all other approaches. 
On GDA test set, similar improvements can also be observed. 
These results demonstrate the effectiveness and generality of our approach.
\subsection{Ablation Study}
\begin{table}[t]
	\centering
	\small
	\begin{tabular}{lcc}
		\toprule
		
		Model & Ign $F_1$  &  $F_1$  	\\ \midrule 
		\textbf{CGM2IR-BERT$_{base}$}  & \textbf{60.02}  & \textbf{62.01}  \\ \midrule
		\textit{w/o} mention integration module & 59.64 & 61.63   \\ 
		\textit{w/o} inter-pair reasoning module  & 59.87 & 61.74  \\  
		\textit{w/o} both module & 59.12 & 60.89  \\ \bottomrule

	\end{tabular}
	\caption{Ablation study of CGM2IR on the development set of DocRED, where ``w/o'' indicates without.}
	\label{tab:result_ablation}
\end{table}

We also conduct a thorough ablation study as shown in Table \ref{tab:result_ablation} to study the contribution of two key modules: context guided mention integration module and inter-pair reasoning module.
From Table \ref{tab:result_ablation}, we can observe that:

(1) When the context guided mention integration module is discarded and replaced with the logsumexp pooling layer, the performance of our model on the DocRED dev set drops by $0.38\%$ in both $F_1$ and Ign $F_1$ score.
Similarly, removal of the inter-pair reasoning module results in a $0.27\%$ drop in $F_1$ and $0.14\%$ in Ign $F_1$.
This phenomenon indicates the effectiveness of context guided mention integration module and inter-pair reasoning module.

(2) Removal of both modules leads to a more considerable decrease.
The $F_1$ score decreases from 62.01\% to 60.89\% and the Ign $F_1$ score decreases from 60.02\% to 59.12\%.
This study demonstrates that all components work together in synergy with the final relation classification. 

\subsection{Intra- and Inter-sentence Triplet Extraction}
To further evaluate the performance, we report the results of intra- and inter-sentence relation extraction on CDR and GDA, since they explicitly annotate these two types of facts.
The experimental results are listed in Table \ref{tab:result_cdrgda}, from which we can find that CGM2IR outperforms the current best models on these two datasets in regard to both intra- and inter-$F_1$.
For example, Our model obtains +5.0 intra-$F_1$/+2.5 inter-$F_1$ and +1.0 intra-$F_1$/+6.1 inter-$F_1$ gain compared with ATLOP on the test set of these two datasets. 
The improvements indicate that our model can effectively capture the complex interactions among entity pairs across the document.
The intra-sentence relations contained in local text can be well considered, as well as the long-distance dependent inter-sentence relations.

\subsection{Effect Analysis for Context Guided Cross-Attention}

To assess the effectiveness of context guided cross-attention in modeling entity representations, we compare five different strategies for generating entity representations including global mean pooling, global max pooling, global attention pooling, global logsumexp pooling, and our context guided cross-attention.
For simplicity, after encoding the document, we directly concat the representations of the head entity and the tail entity then send them to the final classifier.
The results on the development set of DocRED are illustrated in Table \ref{tab:result_mention}, from which we can observe that the context guided cross-attention is absolutely superior to the global strategies.
This result indicates that context guided cross-attention is reasonable and effective, which drives the head and tail entities together to dynamically determine their respective representations.
\subsection{Effect Analysis for Inter-pair Reasoning}
\begin{table}[t]
	\centering
	\small
	\begin{tabular}{lcc}
		\toprule
		
		Method & Ign $F_1$  &  $F_1$  	\\ \hline 
		
		global mean pooling & 57.24 & 58.23   \\ 
		global max pooling  & 57.41 & 58.54  \\ 
		global attention pooling & 58.17 & 59.00  \\ 
		global logsumexp pooling & 58.23 & 59.12  \\ \midrule
		\textbf{context guided cross-attention}  & \textbf{59.34}  & \textbf{60.76}  \\ \bottomrule
		
	\end{tabular}
	\caption{Results of different strategies for generating entity representations on DocRED.}
	\label{tab:result_mention}
\end{table}
\begin{table}[t]
	\centering
	\small
	\begin{tabular}{lccc}
		\toprule
		
		Model & Infer $F_1$  &  $P$  &  $R$	\\ \hline 
		BERT-RE$_{base}^\ast$ & 39.62 & 34.12 & 47.23 \\
		GAIN-GloVe$^\dagger$ & 40.82  & 32.76 & 54.14 \\
		RoBERTa-RE$_{base}^\ast$ & 41.78  & 37.97 & 46.45\\
		SIRE-GloVe$^\dagger$ & 42.72 & 34.83 & 55.22 \\
		GAIN-BERT$_{base}^\ast$ & 46.89 & 38.71 &  59.45  \\ \hline
		\textbf{CGM2IR-BERT$_{base}$}   & \textbf{48.04} & \textbf{39.54} &  \textbf{61.21}  \\ \bottomrule

	\end{tabular}
	\caption{Infer-$F_1$ results on dev set of DocRED. Results with $^\ast$ are reported in \citet{zeng-etal-2020-double}, $^\dagger$ are reported in \citet{zeng-etal-2021-sire}.}
	\label{tab:result_reasoning}
\end{table}
In addition, we evaluate the reasoning ability of our model on the development set of DocRED in Table \ref{tab:result_reasoning}.
Following \citet{zeng-etal-2021-sire}, we use infer-$F_1$ as a metric that only considers instances of the two-hop positive relations in the development set of DocRED.
More specifically, we only evaluate the golden relational facts $r_1, r_2$ and $r_3$ when there exists $e_h\stackrel{r_1}{\longrightarrow}e_o\stackrel{r_2}{\longrightarrow}e_t$ and $e_h\stackrel{r_3}{\longrightarrow}e_t$.

As illustrated in Table \ref{tab:result_reasoning}, CGM2IR outperforms all the baselines in infer-$F_1$.
Specifically, CGM2IR-BERT$_{base}$ improves roughly about $1.15\%$ for infer-$F_1$ score compared with GAIN-BERT$_{base}$.
This reveals that the inter-pair reasoning module plays an important role in capturing intrinsic clues and performing logic reasoning on entities chains.
\section{Conclusion}
In this paper, we propose CGM2IR that incorporates context guided mention integration and inter-pair reasoning to improve DRE.
Instead of simply synthesizing multiple coreferential mentions at once, CGM2IR dynamically generates fine-grained entity representations for each entity pair.
Moreover, we construct a homogeneous entity pair graph and employ GNNs to capture intrinsic clues and perform reasoning among entity pairs.
Experimental results on three widely used DRE datasets demonstrate that our CGM2IR model is effective and outperforms previous state-of-the-art models.

\bibliography{custom}
\bibliographystyle{acl_natbib}

\end{document}